\documentclass[pdflatex,sn-mathphys-num]{sn-jnl}

\usepackage[T1]{fontenc}
\usepackage{graphicx}%
\usepackage{multirow}%
\usepackage{amsmath,amssymb,amsfonts}%
\usepackage{amsthm}%
\usepackage{mathrsfs}%
\usepackage[title]{appendix}%
\usepackage{xcolor}%
\usepackage{textcomp}%
\usepackage{manyfoot}%
\usepackage{booktabs}%
\usepackage{algorithm}%
\usepackage{algorithmicx}%
\usepackage{algpseudocode}%
\usepackage{listings}%
\usepackage{array}
\usepackage{threeparttable}
\usepackage{tabularx}
\usepackage{ragged2e}

\theoremstyle{thmstyleone}%
\theoremstyle{thmstyletwo}%

\theoremstyle{thmstylethree}%

\usepackage{etoolbox}

\begin{document}
\title[Article Title]{VASP Agent: An Agentic Framework for Autonomous First-principles Calculations}


\author[1,2]{\fnm{Zeyu} \sur{Xia}}
\equalcont{These authors contributed equally to this work.}

\author[1,4]{\fnm{Congjie} \sur{Zheng}}
\equalcont{These authors contributed equally to this work.}

\author[1,3]{\fnm{Jinzhe} \sur{Ma}}

\author[1]{\fnm{Zhongyao} \sur{Wang}}

\author[1]{\fnm{Shufei} \sur{Zhang}}

\author[1]{\fnm{Yuqiang} \sur{Li}}

\author[2]{\fnm{Hang} \sur{Su}}

\author[3,6]{\fnm{P.} \sur{Hu}}

\author[4,5]{\fnm{Changshui} \sur{Zhang}}

\author[7]{\fnm{Xingao} \sur{Gong}}

\author[1,8]{\fnm{Wanli} \sur{Ouyang}}

\author[1]{\fnm{Lei} \sur{Bai}}

\author*[1]{\fnm{Dongzhan} \sur{Zhou}}\email{zhoudongzhan@pjlab.org.cn}

\author*[1,9]{\fnm{Mao} \sur{Su}}\email{sumao@pjlab.org.cn}

\affil[1]{\orgname{Shanghai Artificial Intelligence Laboratory}, \orgaddress{\city{Shanghai}, \postcode{200232}, \country{China}}}

\affil[2]{\orgdiv{Department of Computer Science and
Technology}, \orgname{Tsinghua University}, \orgaddress{\city{Beijing}, \postcode{100084}, \country{China}}}

\affil[3]{\orgdiv{School of Physical Science and Technology}, \orgname{ShanghaiTech University}, \orgaddress{\city{Shanghai}, \postcode{201210}, \country{China}}}


\affil[4]{\orgdiv{Department of Automation}, \orgname{Tsinghua University}, \orgaddress{\city{Beijing}, \postcode{100084}, \country{China}}}

\affil[5]{\orgdiv{Beijing National Research Center for Information Science and Technology (BNRist)}, \orgname{Tsinghua University}, \orgaddress{\city{Beijing}, \postcode{100084}, \country{China}}}

\affil[6]{\orgdiv{School of Chemistry and Chemical Engineering}, \orgname{The Queen's University of Belfast}, \orgaddress{\city{Belfast}, \postcode{BT9 5AG}, \country{UK}}}

\affil[7]{\orgdiv{Key Laboratory of Computational Physical Sciences (Ministry of Education), Institute of Computational Physical Sciences, State Key Laboratory of Surface Physics, Department of Physics}, \orgname{Fudan University}, \orgaddress{\city{Shanghai}, \postcode{200433}, \country{China}}}

\affil[8]{\orgname{The Chinese University of Hong Kong}, \orgaddress{\city{Hong Kong}, \postcode{999077}, \country{China}}}

\affil[9]{\orgname{Shenzhen Institute of Advanced Technology, Chinese Academy of Sciences}, \orgaddress{\city{Shenzhen}, \postcode{518055}, \country{China}}}


\abstract{Large Language Models (LLMs) are increasingly embedded in agentic frameworks for scientific discovery. First-principles materials computation imposes a demanding standard for autonomy: successful execution depends on internally consistent inputs, supervision of long-running calculations, and verified outputs.
Here we present VASP Agent, a coding-agent-centered system that combines reusable domain skills, deterministic tools, workspace-state inspection, runtime evidence, and scientific guardrails to execute multi-step VASP calculations.
The system is evaluated across multiple tasks including structural relaxation, bandgap calculation, equilibrium lattice constant determination, and CO/Pt(111) adsorption. VASP Agent completes all evaluated cases, and its computed numerical results are compared with those obtained using pymatgen and other agentic tools. When large deviations occur, the calculation parameters produced by VASP Agent are more appropriate than those produced by LLM-based workflows. Failure analysis shows that errors that terminate fixed pipelines can be diagnosed and recovered under agentic control.}

\maketitle

\section{Introduction}\label{sec1}
The advent of Large Language Models (LLMs) is fundamentally reshaping the landscape of scientific research~\cite{zhang2025exploring, zhang2024comprehensive, zheng2025automation, ai4science2023impact}. Unlike traditional task-specific models, LLMs trained on large-scale corpora of scientific literature and databases can simultaneously perform diverse tasks, including literature summarization, property prediction, and synthesis planning~\cite{zheng2023largelanguagemodelsscientific,ramos2025review}. However, standalone LLMs suffer from critical limitations, including static knowledge, hallucination, and lack of reliable action grounding. Recent agentic systems seek to mitigate these limitations by placing LLMs inside execution harnesses that expose retrieval, structured tool calls, code execution, state management, and logging. Retrieval-Augmented Generation (RAG) can condition model outputs on external or domain-specific information~\cite{lala2023paperqa,huang2024surveyretrievalaugmentedtextgeneration, zhao2024retrievalaugmentedgenerationaigeneratedcontent, prince2024opportunities, gan2025retrieval}, while tool-using agents connect language models to search, computation, and file operations~\cite{bran2023chemcrow,zhang2024honeycomb}. These harnesses improve operational reliability, but scientific computing requires more explicitly defined domain-specific control: actions must be tied to specific input files, runtime evidence, calculation state, and verifiable post-processing.

First-principles materials computation provides a setting in which these requirements become unavoidable. Such calculations offer a controllable and reproducible route for testing hypotheses, exploring parameter spaces, and generating high-fidelity materials data~\cite{Cooper2015}. They also constitute a demanding test of agentic execution, since a calculation is not completed by producing plausible text or issuing a single tool call. A reliable agent for first-principles calculations should assemble consistent input files, preserve dependencies across calculation stages, monitor long-running jobs, distinguish failed from still-running calculations, and extract results only from verified outputs.

Recent progress highlights this potential by using agents to operate specialized scientific software, such as finite-element analysis with MooseAgent~\cite{zhang2025mooseagent}, molecular docking with ChatMol Copilot~\cite{sun2024chatmol}, and hypothesis or research-idea generation~\cite{baek2025researchagentiterativeresearchidea,yang2025moosechemlargelanguagemodels}. Within the computational materials domain, recent multi-agent systems have also been applied to autonomous simulation and scientific-law discovery~\cite{liu2025vaspilotmcpfacilitatedmultiagentintelligence, Han2025PhysAgent}. Nevertheless, systematic evidence from executable first-principles tasks remains limited, and recent AI-for-science benchmarks emphasize that robustness, verifiability, and execution readiness remain open problems~\cite{qin2025scihorizonbenchmarkingaiforsciencereadiness}. This makes it difficult to assess how reliably such agentic harnesses handle calculation setup, input consistency, runtime monitoring, failure recovery, and result extraction in live computational workspaces.

In this work, we introduce VASP Agent, a domain-informed coding agent for multi-step VASP calculations~\cite{kresse1996efficient}. The central contribution is an agentic control layer that integrates reusable domain skills, deterministic tools, stateful workspace inspection, runtime evidence, and scientific guardrails to prepare inputs, execute calculations, monitor progress, diagnose failures, and recover when appropriate. This design enables the agent to make operational decisions that remain physically and procedurally consistent across interdependent first-principles calculation stages. We evaluate the system on representative VASP task families, including structural relaxation, bandgap calculation, equilibrium lattice constant determination, and CO/Pt(111) adsorption. The results show that VASP Agent achieves higher task completion than the LLM-driven workflow baselines and the observed numerical deviations are traceable to material-specific modeling choices.
By demonstrating explicit workspace inspection, runtime monitoring, and recovery across multi-step first-principles calculations, VASP Agent provides a practical reference point for developing more reliable systems for executable scientific workflows.

\section{Results}\label{sec2}

\subsection{Agent framework}
\label{sec:agent-framework}

\begin{figure*}[t]
  \centering
  \includegraphics[width=1.00\textwidth]{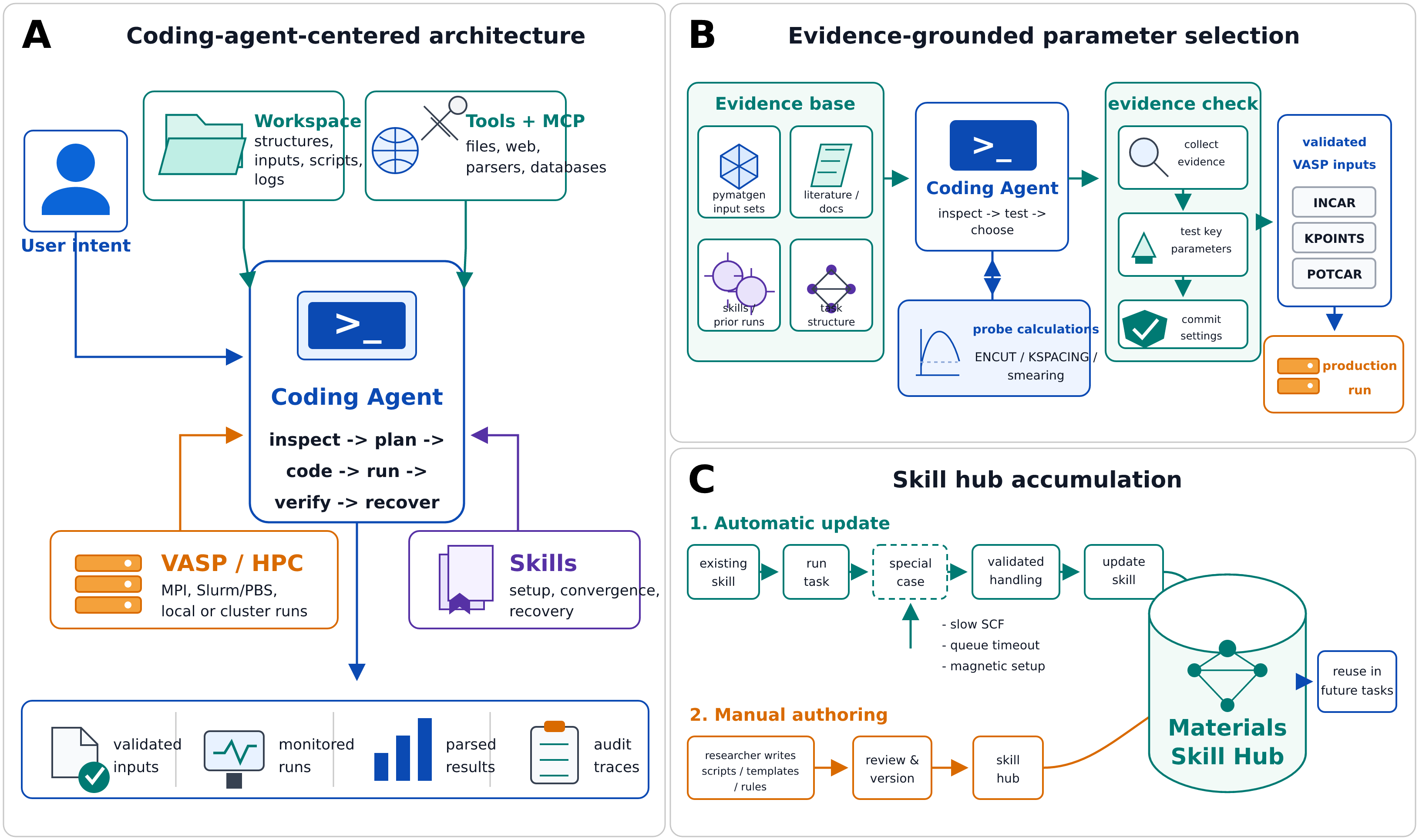}
  \vspace{-1em}
\caption{Overview of VASP Agent.  \textbf{a}, Architecture: a coding agent coordinates the workspace, tools and MCP connectors, VASP/HPC execution resources, and reusable materials-computation skills.  \textbf{b}, Evidence-grounded VASP input-preparation guardrails: the agent combines input-set conventions, literature or documentation evidence, prior skills or runs, task structure, and optional probe calculations to test and select key settings before producing validated VASP inputs and a rationale for the production run.  \textbf{c}, Skill-library accumulation: repository-level skills can be produced either by encoding validated special-case handling from successful runs or by manual authoring from domain users.}
  \label{fig:agent-framework-overview}
\end{figure*}

VASP Agent uses a stateful coding agent as the control plane for the workspace, tools, execution resources, and domain skills. The agent revises its plan in response to logs, intermediate outputs, external evidence, and runtime failures. This design is summarized in Figure~\ref{fig:agent-framework-overview}.  Architecturally, VASP Agent integrates four capability planes (Figure~\ref{fig:agent-framework-overview}a): a persistent project workspace, model context protocol (MCP)-connected tools for external information and computation control, local or high-performance computing (HPC)-based VASP execution, and an explicit skill library.  MCP provides a standard interface for connecting agents to external data sources and tools \cite{AnthropicMCP}, while skills store inspectable instructions and troubleshooting knowledge consistent with skill mechanisms in modern coding agents\cite{AnthropicSkills,OpenAICodexApp}.  These components allow the system to adapt during a calculation instead of executing a predetermined pipeline.

This autonomy is constrained by scientific and operational guardrails for VASP input preparation.  The agent grounds key settings in input-set conventions, literature or documentation evidence, prior skills or runs, task structure, and, when needed, probe calculations (Figure~\ref{fig:agent-framework-overview}b).  The resulting \texttt{INCAR}, \texttt{KPOINTS}, \texttt{POTCAR}, and rationale are validated before production execution.  During run time, this evidence-bound procedure extends to monitoring and recovery through \texttt{OUTCAR}/\texttt{OSZICAR} and run-state records.

\begin{figure*}[t]
\centering
\includegraphics[width=0.98\textwidth]{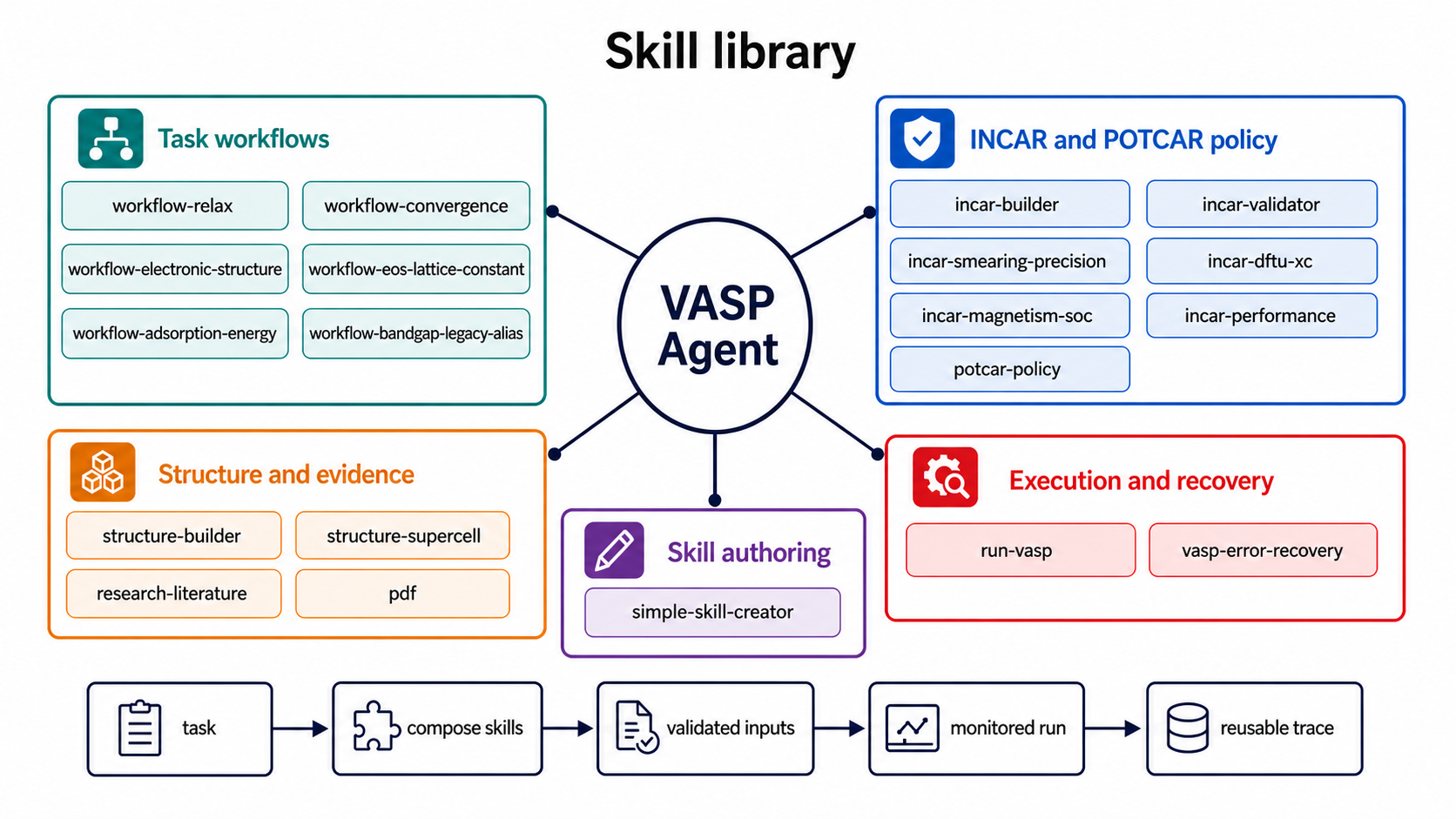}
\caption{Skill library of VASP Agent.  Skills are stored as reviewable repository-level packages and composed during a task rather than hard-coded into a single pipeline.  Name prefixes indicate their primary roles: workflow protocols, INCAR/POTCAR input policy, structure and evidence handling, execution and recovery, and skill authoring.}
\label{fig:vasp-agent-skills}
\end{figure*}

The framework is also designed to accumulate reusable experience, consistent with recent work on cumulative skill creation for scientific agents~\cite{Huang2025CASCADE}.  Skills are repository-level assets that can be reviewed and merged like ordinary code.  The current library contains 20 one-level skill packages whose names expose their primary role: \texttt{workflow-*} skills define task protocols; \texttt{incar-*} skills, together with \texttt{potcar-policy}, constrain VASP input choices; \texttt{structure-*} skills prepare atomic models; and \texttt{run-vasp}/\texttt{vasp-error-recovery} bind execution and repair to calculation evidence.  Figure~\ref{fig:vasp-agent-skills} summarizes this skill-library organization.  As illustrated in Figure~\ref{fig:agent-framework-overview}c, skills can be expanded by encoding validated special-case handling from successful runs or by manual authoring from domain users.  In both cases, successful calculations are converted into reusable procedural knowledge.

Table~\ref{tab:agent-comparison} positions VASP Agent relative to existing workflow frameworks and recent agentic systems.  Traditional frameworks such as pymatgen input sets, atomate/FireWorks, and AiiDA are most effective when the computational protocol has already been specified and the main requirements are high-throughput execution or formal provenance tracking~\cite{Ong2013Pymatgen,Mathew2017Atomate,Pizzi2020AiiDA}.  Recent materials agents increasingly expose multiple domain interfaces and provide partial autonomy, but adapting them to substantially new tasks often still requires manual construction of task-specific workflows~\cite{Chaudhari2026MatSciAgent,Wang2025DREAMS,Liu2025VASPilot,Hu2026TritonDFT}.  Instead of assuming either a fully encoded high-throughput workflow or a fixed set of tool calls, VASP Agent is designed for partially specified, interactive VASP calculation tasks in which the full procedure may not be known in advance. In this setting, the agent operates within a live calculation workspace. It inspects and edits files and scripts, composes tools and skills, responds to runtime failures, and preserves an auditable calculation trace.  A more detailed per-system comparison is provided in Supplementary Table~S1.

\begin{table*}[!t]
\centering
\small
\setlength{\tabcolsep}{5pt}
\renewcommand{\arraystretch}{1.12}
\caption{Position of VASP Agent relative to representative workflow and agent systems.}
\label{tab:agent-comparison}
\begin{tabularx}{\textwidth}{
  @{}>{\raggedright\arraybackslash}p{0.30\textwidth}
  >{\raggedright\arraybackslash}X
  >{\raggedright\arraybackslash}X@{}}
\toprule
\textbf{System class and examples} & \textbf{Automation model} & \textbf{Adaptation to new tasks} \\
\midrule
\textbf{Coded workflow frameworks}\newline
pymatgen input sets~\cite{Ong2013Pymatgen}; atomate/FireWorks~\cite{Mathew2017Atomate}; AiiDA~\cite{Pizzi2020AiiDA}
& Robust execution of specified protocols through libraries, workflow engines, or provenance infrastructure
& New tasks require users to encode protocols as Python recipes, workflows, workchains, or plugins \\
\addlinespace
\textbf{Interface-oriented materials agents}\newline
MatSciAgent~\cite{Chaudhari2026MatSciAgent}; DREAMS / material\_agent~\cite{Wang2025DREAMS}; VASPilot~\cite{Liu2025VASPilot}; TritonDFT~\cite{Hu2026TritonDFT}
& LLM agents connected to domain tools, scientific workflows, or execution interfaces
& Partial autonomy within configured interfaces; substantially new tasks often still require task-specific tools or workflow/interface code \\
\addlinespace
\textbf{VASP Agent}\newline
This work
& Coding-agent-centered VASP workspace with tools, HPC execution, monitoring, recovery, and reusable skills
& Adapts by inspecting files, editing scripts, composing tools, verifying runtime evidence, and accumulating reusable task skills \\
\bottomrule
\end{tabularx}
\end{table*}

\subsection{VASP task evaluation}


Task completion and task-specific numerical metrics are selected as evaluation criteria (see Methods).
VASP Agent employs DeepSeek-V3.2~\cite{deepseekai2025deepseekv32} as the backbone.
For the structural relaxation and bandgap calculation tasks, we constructed an LLM-powered workflow baseline to disentangle the contribution of the base LLM from that of the surrounding agent harness. In this baseline, the computational procedure was predefined, and the LLM was tasked only with recommending the INCAR parameters, without iterative execution, error recovery, or result-driven refinement. We evaluated this setting using DeepSeek-V3.2~\cite{deepseekai2025deepseekv32}, Gemini-3.1-Pro Preview~\cite{google2026gemini31propreview}, GPT-5.2~\cite{openai2025gpt52systemcard}, and Qwen3.5-397B-A17B~\cite{qwen2026qwen35}.
In addition, INCAR settings recommended by pymatgen were used as a reference for comparison. These pymatgen-generated parameters should be viewed as reference settings rather than ground-truth parameter choices.
For cases exhibiting substantial discrepancies in the computed results, we analyzed the differences among the methods in their selection of INCAR parameters.
For the equilibrium lattice constant and adsorption energy tasks, we adopted the settings used in DREAMS~\cite{Wang2025DREAMS}. In particular, the Sol27LC dataset~\cite{Wellendorff2012BEEFvdW} used for the lattice constant task contains results calculated by experts, and these values were treated as the ground-truth for comparison.

\begin{table*}[htbp]
    \centering
    \small
    \renewcommand{\arraystretch}{1.12}
    \setlength{\tabcolsep}{3pt}
    \begin{tabularx}{\textwidth}{
        >{\raggedright\arraybackslash}p{0.12\textwidth}
        >{\raggedright\arraybackslash}p{0.23\textwidth}
        >{\centering\arraybackslash}p{0.10\textwidth}
        >{\centering\arraybackslash}p{0.10\textwidth}
        >{\centering\arraybackslash}p{0.13\textwidth}
        >{\centering\arraybackslash}p{0.11\textwidth}
        >{\centering\arraybackslash}p{0.11\textwidth}}
        \toprule
        \textbf{System} & \textbf{Model backbone} &
        \textbf{Completion} & $\overline{d}_E$ &
        $\widetilde{d}_E$ & $\overline{d}_R$ &
        $\widetilde{d}_R$ \\
        \midrule
        \multirow{4}{=}{Workflow baseline} & DeepSeek-V3.2 &
        36/40 & 80.36 & 2.20 & 0.0234 & 0.0050 \\
        & Gemini-3.1-Pro &
        35/40 & 45.48 & 2.86 & 0.0152 & 0.0041 \\
        & GPT-5.2 &
        31/40 & 81.37 & 5.61 & 0.0318 & 0.0149 \\
        & Qwen3.5-397B-\allowbreak A17B &
        36/40 & 82.63 & 3.21 & 0.0154 & 0.0051 \\
        \addlinespace
        VASP Agent & DeepSeek-V3.2 &
        40/40 & 62.42 & 5.65 & 0.0218 & 0.0051 \\
        \bottomrule
    \end{tabularx}
\caption{
Results for the structural relaxation task. The energy deviation metric mean $\overline{d}_E$ and median $\widetilde{d}_{E}$ are reported in meV/atom, and the normalized root mean square deviation (RMSD) metric mean $\overline{d}_R$ and median $\widetilde{d}_{R}$ are dimensionless.
The computational settings recommended by pymatgen rather than human experts are used as reference values.
Completion denotes the number of systems with matched final structures in the RMSD evaluation.
The metrics were calculated only for cases in which the computations were successfully completed.
}
    \label{tab:main_results_sr}
\end{table*}

\begin{table*}[htbp]
    \centering
    \small
    \renewcommand{\arraystretch}{1.12}
    \setlength{\tabcolsep}{5pt}
    \begin{tabularx}{\textwidth}{
        >{\raggedright\arraybackslash}p{0.21\textwidth}
        >{\raggedright\arraybackslash}p{0.23\textwidth}
        >{\centering\arraybackslash}p{0.10\textwidth}
        >{\centering\arraybackslash}p{0.13\textwidth}
        >{\centering\arraybackslash}p{0.16\textwidth}}
        \toprule
        \textbf{System} & \textbf{Model backbone} &
        \textbf{Completion} & $\overline{d}_{E_g}$ & $\widetilde{d}_{E_g}$ \\
        \midrule
        \multirow{4}{=}{Workflow baseline} & DeepSeek-V3.2 &
        21/24 & 0.099 & 0.066 \\
        & Gemini-3.1-Pro &
        24/24 & 0.101 & 0.073 \\
        & GPT-5.2 &
        21/24 & 0.130 & 0.081 \\
        & Qwen3.5-397B-\allowbreak A17B &
        24/24 & 0.105 & 0.073 \\
        \addlinespace
        VASP Agent & DeepSeek-V3.2 &
        24/24 & 0.159 & 0.088 \\
        \bottomrule
    \end{tabularx}
\caption{
Results for the bandgap calculation task. The bandgap deviations Mean $\overline{d}_{E_g}$ and median $\widetilde{d}_{E_g}$ are reported in eV.
The computational settings recommended by pymatgen rather than human experts are used as reference values.
Completion denotes the number of systems with successfully parsed final band gaps.
The metrics were calculated only for cases in which the computations were successfully completed.
}
    \label{tab:main_results_bg}
\end{table*}

\begin{table*}[htbp]
    \centering
    \small
    \renewcommand{\arraystretch}{1.12}
    \setlength{\tabcolsep}{5pt}
    \begin{tabularx}{\textwidth}{
        >{\raggedright\arraybackslash}X
        >{\centering\arraybackslash}p{0.14\textwidth}
        >{\centering\arraybackslash}p{0.18\textwidth}
        >{\centering\arraybackslash}p{0.18\textwidth}}
        \toprule
        \textbf{System} & \textbf{Completion} & $\overline{d}_{a}$ &
        $\widetilde{d}_{a}$ \\
        \midrule
        DREAMS & 27/27 & 0.526 & 0.542 \\
        VASP Agent & 27/27 & \textbf{0.374} & \textbf{0.258} \\
        \bottomrule
    \end{tabularx}
\caption{Results for the Sol27LC equilibrium lattice constant task. Completion denotes the number of systems for which the equation-of-state workflow successfully returns an equilibrium lattice constant. The metric mean $\overline{d}_{a}$ and median $\widetilde{d}_{a}$ measure the relative deviation of the agent-calculated lattice constant from the human-expert calculated value~\cite{Wellendorff2012BEEFvdW}, and are reported as percentages. The DREAMS reference results are taken from Ref.~\cite{Wang2025DREAMS}. Best results are shown in bold.}
    \label{tab:main_results_lc}
\end{table*}

\begin{table*}[htbp]
    \centering
    \small
    \renewcommand{\arraystretch}{1.12}
    \setlength{\tabcolsep}{5pt}
    \begin{tabularx}{\textwidth}{
        >{\raggedright\arraybackslash}p{0.22\textwidth}
        >{\raggedright\arraybackslash}p{0.22\textwidth}
        >{\centering\arraybackslash}p{0.18\textwidth}
        >{\centering\arraybackslash}p{0.18\textwidth}}
        \toprule
        \textbf{Adsorption site} & \textbf{Orientation} &
        \textbf{$E_{\mathrm{ads}}$} & \textbf{$\Delta E_{\mathrm{ads}}$} \\
        \midrule
        fcc & upright &
        -1.756 & 0.000 \\
        fcc & tilted-$x$ &
        -1.752 & 0.004 \\
        fcc & tilted-$y$ &
        -1.752 & 0.004 \\
        ontop & upright &
        -1.591 & 0.165 \\
        ontop & tilted-$x$ &
        -1.586 & 0.169 \\
        ontop & tilted-$y$ &
        -1.587 & 0.169 \\
        \bottomrule
    \end{tabularx}
\caption{Adsorption energies for the CO/Pt(111) configurations evaluated by VASP Agent. $E_{\mathrm{ads}}$ and $\Delta E_{\mathrm{ads}}$ are reported in eV. $E_{\mathrm{ads}}=E_{\mathrm{adsorbed}}-E_{\mathrm{surface}}-E_{\mathrm{CO}}$. $\Delta E_{\mathrm{ads}}$ is computed relative to the most stable configuration, fcc upright.}
    \label{tab:main_results_adsorption}
\end{table*}

The evaluation metrics are summarized in Tables~\ref{tab:main_results_sr}--\ref{tab:main_results_adsorption}. In the structural relaxation and bandgap calculation tasks, VASP Agent completes all cases, whereas some baseline runs terminate before producing valid outputs.
The median values of the selected metrics were small, whereas the corresponding mean values were substantially larger. This distribution indicates that, for most cases, the computational results differed only slightly, while a small number of cases exhibited pronounced discrepancies.
Several larger deviations are explained by material-specific settings selected by VASP Agent: DFT+$U$ for Ti-containing oxides~\cite{dudarev1998electron}, D3(BJ) and van der Waals corrections for layered tellurides~\cite{grimme2010consistent,grimme2011effect,bjorkman2012van}, and denser K-point sampling or higher \texttt{ENCUT} for selected bandgap systems.
For these cases, the computational settings recommended by VASP Agent are more appropriate than those recommended by pymatgen.

For the equilibrium lattice constant task, the expert-calculated lattice constants are available, and both DREAMS and VASP Agent successfully completed all 27 systems. The comparison therefore primarily assesses the accuracy of the equation-of-state calculations. 
VASP Agent achieves lower mean and median relative deviations than DREAMS~\cite{Wang2025DREAMS}, and this can be attributed to the design of VASP Agent, which validates the physical consistency of the equation-of-state curve before determining the equilibrium lattice constant, requiring a smooth energy--volume relation that brackets the minimum.

The CO/Pt(111) task examines a different dimension of computational reliability. In this case, the agent calculates the gas-phase molecule, the clean slab, and multiple adsorbed slab configurations, and then evaluates adsorption energies under a consistent reference convention.
VASP Agent identifies fcc upright as the most stable configuration. The two tilted fcc structures lies only about 0.004 eV higher in adsorption energy, whereas the lowest-energy ontop configuration is higher by 0.165 eV. This preference for fcc adsorption over ontop adsorption falls within the range of literature results summarized by DREAMS~\cite{Wang2025DREAMS}, and therefore supports the consistency of the multi-configuration workflow.

\noindent\textbf{Mechanisms of reliable execution.} These outcomes are enabled by three operational controls: evidence-grounded VASP input preparation, multi-stage state management, and runtime verification. During input preparation, VASP Agent composes workflow-level and INCAR-policy skills to select \texttt{INCAR} parameters such as cutoff, sampling, smearing, relaxation controls, and stage-specific HSE tags from task context, local evidence, and retrieved reference information (see Supplementary S1). When convergence settings are uncertain, the agent can vary \texttt{ENCUT} and \texttt{KSPACING} for the target system to identify settings that balance numerical accuracy and computational cost. For energy-difference calculations, it keeps comparable settings, including pseudopotentials, cutoff, sampling, and smearing, consistent across the relevant subcalculations.

At the task level, these controls take different forms. In structural relaxations, the agent builds inputs from the crystal structure, selects relaxation settings suited to the material, keeps the calculation files consistent, and uses recorded state to decide subsequent actions. Bandgap calculations require state to carry across stages: the agent preserves compatible \texttt{ENCUT}/\texttt{KSPACING} choices, \texttt{WAVECAR}/\texttt{CHGCAR} continuation files, and HSE-specific tags from the PBE stage to the HSE stage before extracting the final bandgap. For lattice constants, the agent obtains elemental structures, generates scaled cells, verifies each static calculation, fits the equation of state, and checks the fit before reporting the equilibrium value. Adsorption calculations require systematic enumeration of candidate sites and consistent energy accounting for the gas molecule, clean slab, and adsorbed slabs. Supplementary Table~S6 provides a parameter-level LiFePO$_4$ relaxation example, showing how VASP Agent preserves key \texttt{MPRelaxSet}/\texttt{MITRelaxSet} choices while adding structure-specific validation and runtime recovery controls.

\begin{figure*}[htbp]
    \centering
    \includegraphics[width=0.98\textwidth]{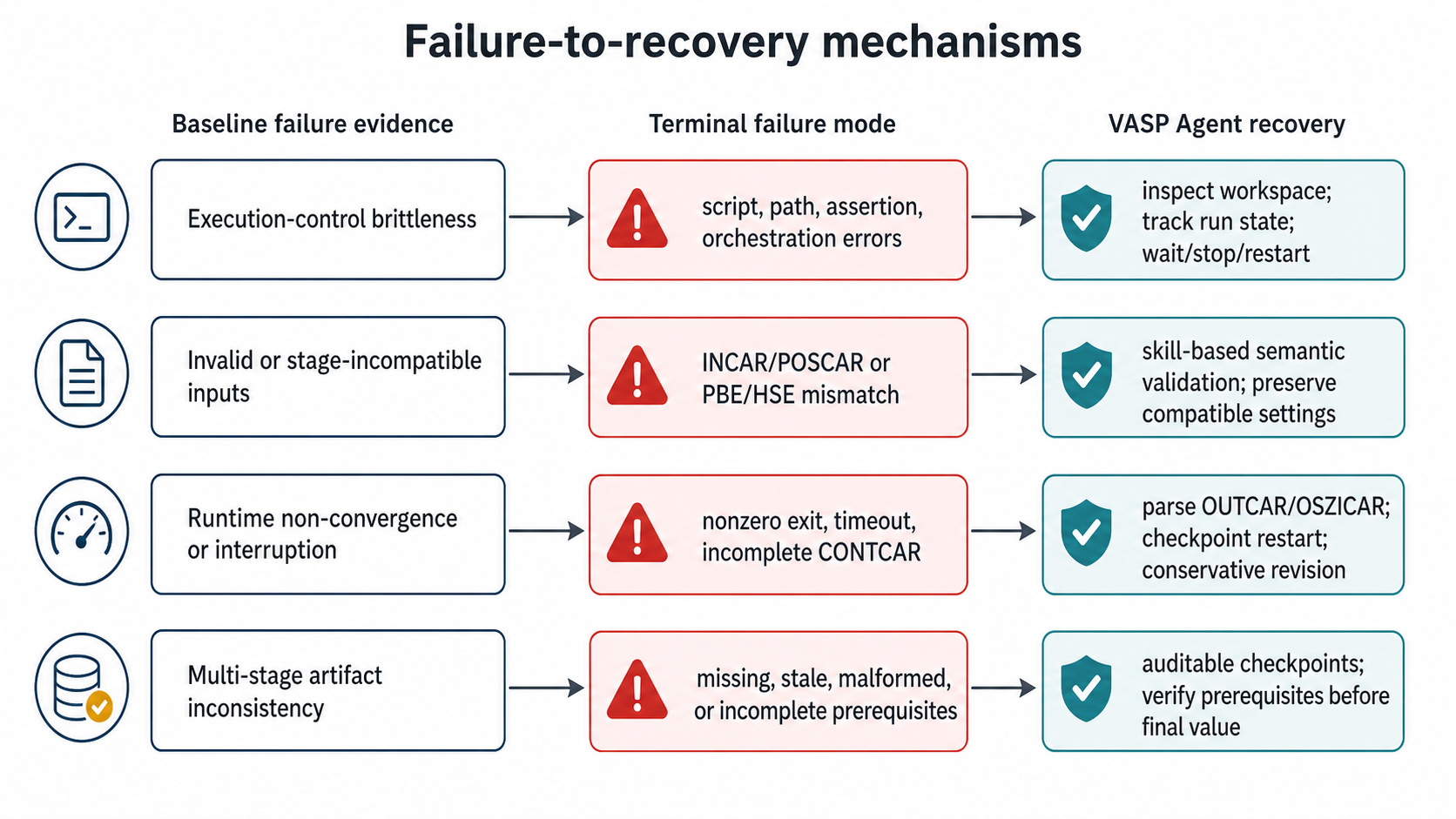}
\caption{Common workflow failure patterns and the corresponding recovery mechanisms in VASP Agent. These failures are primarily failures of execution control, semantic validation, runtime monitoring, or multi-stage state management.}
    \label{fig:fixed_failure_patterns}
\end{figure*}

\noindent\textbf{Workflow Failures and Agent Recovery}
In the present evaluation, VASP Agent completes all tested cases.  The terminal failures analyzed here therefore arise from the workflow baselines. Workflow baselines usually treat malformed input or a missing downstream output as an unrecoverable failure, whereas VASP Agent can treat these events as intermediate states and try to repair the relevant inputs and rerun the calculation when the evidence supports doing so. Figure~\ref{fig:fixed_failure_patterns} summarizes the common failure patterns observed in the workflow baselines and the corresponding recovery mechanisms used by the agent; a representative LiFePO$_4$ relaxation case in Supplementary Fig.~S1 shows how runtime evidence, skill-guided input revision, and state-aware reruns recover from electronic non-convergence to a verified endpoint.

\section{Discussion}\label{sec3}
The pursuit of end-to-end scientific discovery, which integrates hypothesis generation, experimental design, execution, and results validation into an automated workflow, is a central challenge in AI for science. This work takes a step forward in materials computation by introducing an LLM-powered coding agent that can autonomously perform first-principles calculations reliably.

Realistic computational materials tasks often involve multiple interdependent steps with strict sequencing, such that the outcome of one step determines the inputs or constraints of the next. These dependencies place higher demands on an agent than on a prompt-only model or a fixed script. The results indicate that three capabilities are especially important: domain skills and tools, especially INCAR/POTCAR policy and structure-building skills, that constrain VASP input preparation; multi-step state management that keeps structures, parameters, and intermediate outputs consistent across stages; and runtime verification and recovery based on calculation evidence. These capabilities explain why VASP Agent improves completion over workflow baselines while remaining consistent with the relevant reference protocols. Differences from the pymatgen reference are traceable to explicit modeling choices rather than unexplained variation.

The broader implication is that scientific agents should be evaluated not only by final scalar errors but also by the quality of their executable decision traces. Once a mature DFT protocol is fixed, different correct implementations often produce similar numerical values. The harder problem is whether the agent can construct the protocol, keep it consistent, detect failures, and leave an auditable trail. VASP Agent is designed around this requirement, providing a practical route toward automated computational experimentation.

\section{Methods}\label{sec4}

\subsection{Dataset and Evaluation Metrics}\label{subsec:dataset_metrics}

All density functional theory (DFT) calculations were performed using the Vienna Ab initio Simulation Package (VASP)~\cite{kresse1996efficient} with projector-augmented-wave potentials~\cite{Blochl1994PAW,Kresse1999PAW}. The benchmark protocols use semilocal exchange-correlation functionals~\cite{Perdew1996PBE}, optional DFT+$U$ corrections~\cite{dudarev1998electron}, dispersion corrections where appropriate~\cite{grimme2010consistent,grimme2011effect}, and HSE hybrid-functional stages for the bandgap task~\cite{Heyd2003HSE,Krukau2006HSE}. Each system was assessed according to two criteria: workflow completion, defined as the successful production of a parseable output from the required computational procedure, and a task-specific numerical metric. For structural relaxation and bandgap calculation, the numerical metrics measure consistency with reference calculations generated from pymatgen input sets. For equilibrium lattice constants and CO/Pt(111) adsorption, the metrics follow the settings in DREAMS~\cite{Wang2025DREAMS}.

For tasks compared against a reference calculation, we first compute a per-system absolute deviation $d_i(x)$ for the target quantity $x$. Over the completed comparison set $\mathcal{C}_x$, we report the mean and median deviations,
\[
\overline{d}_x =
\frac{1}{|\mathcal{C}_x|}\sum_{i\in \mathcal{C}_x} d_i(x),
\quad
\widetilde{d}_x =
\operatorname{median}_{i\in \mathcal{C}_x} d_i(x).
\]

\paragraph{Structural Relaxation.}

The structural relaxation dataset contains 40 geometry-optimization tasks. Initial structures were obtained from the Materials Project database, and the reference calculations were generated from pymatgen input sets~\cite{Ong2013Pymatgen}. The dataset spans metallic elements and alloys, energy-storage materials, functional oxides, and semiconductors. To enhance structural complexity, each system contains more than 25 atoms and includes point defects; for materials with small primitive cells, supercell expansion was performed before introducing defects.

Completion is reported as the number of systems whose final structure can be parsed and matched to the reference final structure. Numerical consistency is measured by both energy and structural agreement. For each system with a parsed energy, the energy deviation per atom is
\[
d_i(E) =
\frac{
\left|E_i^{\text{pred}}-E_i^{\text{ref}}\right|
}{N_i},
\]
where $N_i$ is the common atom count in the predicted and reference final structures. Structural agreement is measured after matching the predicted and reference final structures with a $30^\circ$ lattice-angle tolerance. For a matched system $i$, the normalized root mean square deviation (RMSD) is
\[
R_i =
\frac{1}{\ell_i}
\left[
\frac{1}{N_i}
\sum_{j=1}^{N_i}
\left\|
\mathbf{r}_{ij}^{\mathrm{pred}}
-
\mathbf{r}_{i\pi_i(j)}^{\mathrm{ref}}
-
\mathbf{t}_i
\right\|^2
\right]^{1/2},
\quad
\ell_i=\left(\frac{V_i}{N_i}\right)^{1/3},
\]
where $N_i$ again denotes the common atom count in the predicted and reference final structures, $\pi_i$ is the site mapping, $\mathbf{t}_i$ is the optimal global translation, and $V_i$ is the matched-cell volume used to normalize the RMS displacement by the characteristic volume per atom. Final structures are read from \texttt{CONTCAR}, with the final structure in \texttt{vasprun.xml} used as a fallback when needed. Table~\ref{tab:main_results_sr} reports $\overline{d}_E$, $\widetilde{d}_E$, and the mean and median of $R_i$ over matched structures.

\paragraph{Bandgap Calculation.}

The bandgap dataset contains 24 semiconductor tasks. Initial structures were obtained from the Springer Materials database~\cite{SpringerMaterials}. The materials cover narrow to ultra-wide band gaps and include both direct and indirect semiconductors, including representative systems such as Si and ZnO. The target workflow is a multi-step PBE-to-HSE electronic structure calculation, and completion requires a parseable band gap from the final electronic-structure output. The DFT reference band gaps were generated from pymatgen input sets under the same target protocol. For each parsed band gap, the absolute deviation is
\[
d_i(E_g) =
\left|E_{g,i}^{\text{pred}}-E_{g,i}^{\text{ref}}\right|,
\]
where $\mathcal{C}_{E_g}$ is the set of bandgap calculation tasks with parseable predicted band gaps. Table~\ref{tab:main_results_bg} reports $\overline{d}_{E_g}$ and $\widetilde{d}_{E_g}$ over $\mathcal{C}_{E_g}$.

\paragraph{Equilibrium Lattice Constant.}

The lattice constant dataset follows the Sol27LC setting~\cite{Wellendorff2012BEEFvdW} adopted in DREAMS~\cite{Wang2025DREAMS}. It contains 27 elemental crystals covering BCC, FCC, and diamond structures. Each task starts from a natural-language request specifying the element and crystal type. Completion requires all static energy calculations to finish, the equation of state to be fitted, and an equilibrium lattice constant to be returned. Accuracy is measured against the human-expert DFT lattice constants reported in the Sol27LC dataset using the relative lattice-constant deviation,
\[
d_i(a) =
\frac{\left|a_i^{\text{pred}}-a_i^{\text{ref}}\right|}
     {\left|a_i^{\text{ref}}\right|}.
\]
Here $\mathcal{C}_a$ is the set of systems for which an equilibrium lattice constant is returned. Table~\ref{tab:main_results_lc} reports $\overline{d}_a$ and $\widetilde{d}_a$ over $\mathcal{C}_a$; in the completed VASP Agent and DREAMS-reference runs reported here, $|\mathcal{C}_a|=27$.

\paragraph{CO/Pt(111) Adsorption.}

The adsorption task follows the CO/Pt(111) setting in DREAMS~\cite{Wang2025DREAMS}, including its literature-grounded reference context for the CO/Pt(111) site-preference problem. 
VASP Agent constructs a gas-phase CO molecule, a 4-layer p(2$\times$2) Pt(111) slab with a 15~\AA{} vacuum region, and six CO/Pt(111) initial configurations: fcc and ontop sites, each with upright, tilted-$x$, and tilted-$y$ orientations.  The enumerated structures target the fcc and ontop site families used for the reference comparison.  All adsorbed structures use C-down anchoring and are validated before execution.
Completion requires convergence of the gas-phase CO, clean slab, and all enumerated adsorbed configurations, followed by successful adsorption-energy post-processing. For each adsorbed configuration $j$,
\[
E_{\mathrm{ads}}^{(j)} =
E_{\mathrm{adsorbed}}^{(j)}-E_{\mathrm{surface}}-E_{\mathrm{CO}},
\]
and the relative adsorption energy reported in Table~\ref{tab:main_results_adsorption} is
\[
\Delta E_{\mathrm{ads}}^{(j)} =
E_{\mathrm{ads}}^{(j)}-\min_k E_{\mathrm{ads}}^{(k)}.
\]
The site-preference energy is computed as
\[
\Delta E_{\mathrm{site}} =
\min_{j\in\mathrm{ontop}}E_{\mathrm{ads}}^{(j)}
-
\min_{j\in\mathrm{fcc}}E_{\mathrm{ads}}^{(j)}.
\]
The per-configuration raw energies, adsorption energies, and ionic steps are reported in Supplementary Table~S11.

\subsection{Workflow Baseline Implementation}\label{subsec:workflow_baseline}

For the structural relaxation and bandgap tasks, the reference calculations generated from pymatgen input sets define the numerical comparison protocol. The workflow baseline is separate: it is implemented as a modular but fixed execution harness built around pymatgen-style VASP input preparation~\cite{Ong2013Pymatgen}. The evaluation input directories and deterministic setup logic use existing programmatic interfaces for structures, pseudopotentials, K-point generation, and parsing. In particular, the pseudopotential choices follow pymatgen/Materials Project recommendations encoded in \texttt{MPRelaxSet}. The LLM is used only for selected parameter-generation steps, mainly the generation of the requested INCAR-like text artifact. Each task is represented by a predefined sequence of pipeline components. The task metadata file specifies the natural-language instruction, output filename, extraction pattern, and shell command for each step, after which the task class executes the corresponding component chain in order. These components include prompt construction, an LLM call, regular-expression extraction of the generated VASP input, file writing, deterministic shell-command execution, and final result extraction. In this design, the LLM supplies only the requested textual artifact, such as an INCAR file, while file operations and VASP/post-processing commands are executed by deterministic Python components.

This baseline should therefore be understood as a controlled workflow system. It can copy the dataset input directory, write generated files, run the prescribed command, and invoke predefined parsers such as energy or bandgap extraction scripts. However, its operation is restricted to predefined programmatic interfaces and does not invoke the open-ended agent loop shown in Figure~\ref{fig:agent-framework-overview}. In terms of execution guardrails, the baseline lacks the evidence-bound checks used by VASP Agent during live calculations. It does not bind actions to a verified run directory or job state, nor does it use evidence from OUTCAR/OSZICAR or scheduler state to decide whether outputs should be accepted. It also lacks the broader agent loop for pre-run information gathering from literature, web pages, local prior calculations, or reusable skills. Once a predefined command has returned or timed out, the baseline follows the fixed pipeline rather than diagnosing the cause of failures and revising the workflow. This distinction is central to understanding the completion differences between the baseline and VASP Agent.
\section*{Data Availability}

The benchmark data package supporting this study is deposited on figshare (DOI/link to be inserted before publication).  The package contains the public task inputs, filtered raw calculation evidence, and machine-readable result tables for the structural relaxation, HSE bandgap, Sol27LC lattice-constant, and CO/Pt(111) adsorption benchmarks.  It also includes manifest and data-dictionary files that map each benchmark case to the corresponding inputs, calculation evidence, and summary metrics.  Copyrighted pseudopotential files and large restart or density files are not included.

\section*{Code availability}

The code for reproducing the results in this paper can be found at \url{https://github.com/Phoinikas03/VASP_Agent}.

\section*{Acknowledgments}
This work was supported by the New Generation Artificial Intelligence National Science and Technology Major Project (2025ZD0121802) and the National Natural Science Foundation of China (Grant No. 12404291). Z.X., C.Zheng, and J.M. did this work during their internship at Shanghai Artificial Intelligence Laboratory.

\section*{Author contributions}
W.O., D.Z., and M.S. conceived the idea and designed the research. Z.X., C.Zheng, J.M., and Z.W. developed the agent framework, performed the experiments, and wrote the first draft. S.Z., Y.L., H.S., P.H., C.Zhang, X.G. and L.B. contributed technical ideas. All authors discussed the results and reviewed the manuscript. D.Z. and M.S. supervised the work.

\section*{Competing interests}
The authors declare no competing interests.

\bibliography{sn-bibliography}

\end{document}